\documentclass{article} % For LaTeX2e
\usepackage{iclr2027_conference,times}

\usepackage{amsmath,amsfonts,bm}

\def\eqref#1{equation~\ref{#1}}
\def\1{\bm{1}}

\DeclareMathAlphabet{\mathsfit}{\encodingdefault}{\sfdefault}{m}{sl}
\SetMathAlphabet{\mathsfit}{bold}{\encodingdefault}{\sfdefault}{bx}{n}

\usepackage{hyperref}
\usepackage{graphicx}
\usepackage{url}
\usepackage{booktabs}
\usepackage{booktabs}
\usepackage{mathtools}

\title{Meta Expert: Cross-Layer Expert Sharing for Storage- and Retrieval-Efficient MoE}
\iclrfinalcopy

\author{Ermo Hua$^{1,2,*}$, \ Xiangyu Hong$^{1,2,*}$, \ Baiting Wu$^{1,2,*}$, \ Che Jiang$^{1,*}$\\[4pt]
\textbf{Haozheng Hou$^{2}$, \ Yicheng Gu$^{1,2}$, \ Hanjing Wang$^{2}$, \ Qirui Zeng$^{1}$, \ Ming Zhang$^{2}$} \\[4pt]
\textbf{Ning Ding$^{1,2}$, \ Qipeng Guo$^{2}$, \ Youbang Sun$^{1,2,\dagger}$, \ Biqing Qi$^{2,\dagger}$, \ Bowen Zhou$^{1,2,\dagger}$} \\[4pt]
$^1$Tsinghua University, $^2$Shanghai AI Laboratory
}

\begin{document}

\maketitle

\fancyhead[L]{Preprint}

\begingroup
\renewcommand{\thefootnote}{\fnsymbol{footnote}}
\footnotetext[1]{Equal contribution. $^\dagger$Corresponding author.}
\endgroup

\begin{abstract}
% 随着模型参数量的变大，稀疏激活的 MoE 方案已经成为模型 Scaling 的必需品。然而，现在的主流模型采取分层结构，不同层的 Experts 虽然在拓扑上相互独立，但在存储内容上存在着很大的相似性和冗余性。相比于层级化排布的 Local Expert，我们推测跨层共享的 Meta Expert 机制有更大的潜力。

% 为此，我们对 Local Expert 和 Meta Expert 进行了多维度的对比，发现引入 Meta Expert 之后的模型存在着更低的参数冗余，并有着更低的交叉熵损失、更高的下游任务准确率。在保持激活数量和专家颗粒度不变的情况下，我们的模型仅使用 70% 的专家数量就可以达到与 Vanilla MoE 相近的效果，大大提升了模型的 Scaling 潜力。

% Meta Experts 除了可以去除层间 Expert 的冗余、提高信息存储效率，我们的理论分析还进一步指出，相比于 Local Experts， Meta Experts 还为 hidden states 提供了更加灵活的 Residual Path，提高了信息读取效率。

% 至于 Meta Experts 存储效率更高的成因，我们进一步的理论分析也指出，相比于 Local Experts， Meta Experts 还为 hidden states 提供了更加灵活的 Residual Path，这在原理上避免了不同层存储功能相近的专家。

As model parameters scale up, sparsely activated Mixture-of-Experts (MoE) approaches have become essential for model scaling. However, mainstream models currently adopt hierarchical architectures, where experts in different layers, although topologically independent, exhibit substantial similarity and redundancy in their stored content. Compared to the hierarchically arranged \textbf{Local Expert} in existing mainstream models, we hypothesize that inter-layer shared \textbf{Meta Expert} hold greater potential. 
To this end, we conduct a multi-dimensional empirical comparison between Local and Meta Experts, and find that models incorporating Meta Experts exhibit lower parameter redundancy, together with lower cross-entropy loss and higher downstream task accuracy. With the activated quantity and the expert granularity held fixed, \textbf{our model matches the performance of Vanilla MoE using only 70--80\% of the experts}, substantially improving the model's scaling potential.
Beyond removing inter-layer redundancy among experts and improving information storage efficiency, our theoretical analysis further shows that, compared with Local Experts, Meta Experts provide hidden states with more flexible residual pathways, thereby improving information retrieval efficiency.

\end{abstract}

\section{Introduction}

\begin{figure}[ht]
    \centering
    \includegraphics[width=\textwidth]{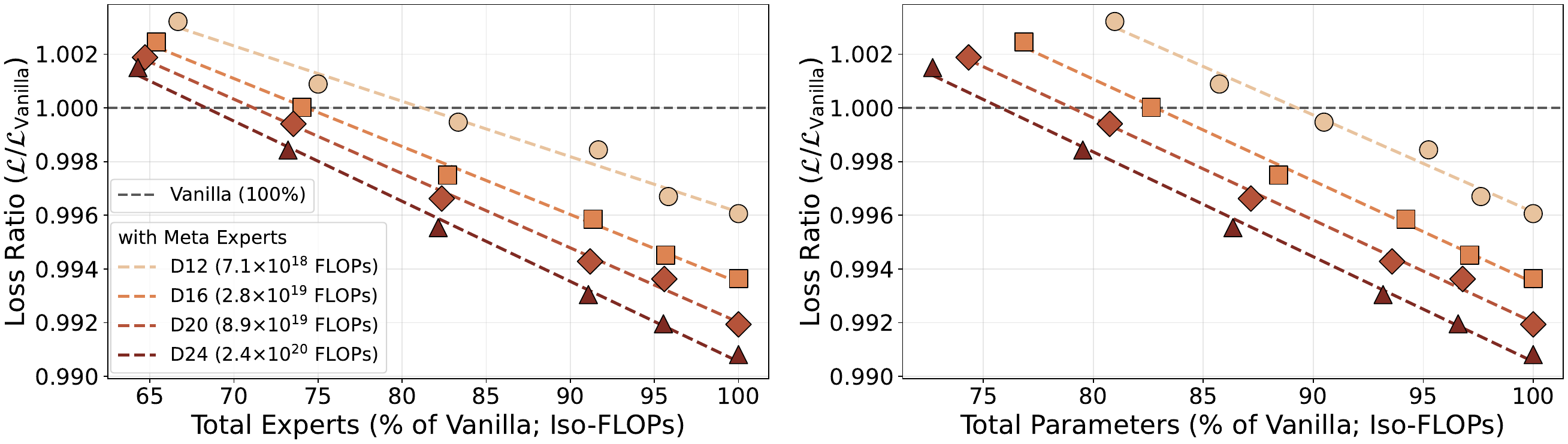} 
    \caption{Comparison between Vanilla and Meta-Expert models. All models are
    trained under the optimal setting of the Vanilla model, at varying depths
    and scales. Models with Meta Experts achieve comparable loss to the Vanilla
    baselines with only 70--80\% as many experts, or 75--90\% as many
    parameters, and the savings in both experts and parameters grow as the
    models get deeper and larger, indicating the favorable scalability of Meta
    Experts.}
    \label{fig:ladder-paritial-experts}
\end{figure}

% 近年来，大语言模型在通用文本、数学、代码等领域的能力正逐渐接近并超过人类。在这个过程中，越来越大规模的模型参数量和训练数据量功不可没，人们发现模型的智能可以随着这两方面的规模化显著提升。

In recent years, large language models (LLMs) have gradually approached or even surpassed human-level ability in general text, mathematics, and code~\citep{fewshotlearners,gpt4}. This progress owes much to the increasing scale of model parameters and training data: intelligence has been improved substantially as both factors grow~\citep{kaplan2020scalinglawsneurallanguage,hoffmann2022trainingcomputeoptimallargelanguage}.

% 为了让模型可以变得越来越大，参数稀疏化成为了重要的途径。作为模型中占据巨大多数比例参数的模块，前馈网络最初的形态是基于稠密线性矩阵运算的。但随着模型参数量的变大，这种稠密运算逐渐变得不可承受，为训练和推理带来了效率瓶颈。同时，虽然前馈网络中采用稠密运算，但是在一个词元前向传播时被真正激活的参数是非常稀疏的，这些没有被真正激活的参数却仍要参与这次计算，产生了计算浪费。

To enable ever-larger models, parameter sparsification has become an important approach~\citep{shazeer2017,switchtransformers,muennighoff2025olmoe}. As the module that accounts for the vast majority of model parameters, the feed-forward network was originally a densely computed block.~\citep{attentionisallyouneed}. However, as model size grows, such dense computation gradually becomes prohibitively expensive, creating efficiency bottlenecks for both training and inference~\citep{lepikhin2021gshard,switchtransformers}. Meanwhile, although feed-forward networks are densely computed, only a sparse subset of parameters is actually activated for a given token during its forward pass~\citep{li2023lazyneuronphenomenonemergence,mirzadeh2024relu}; the remaining inactive parameters still participate in the computation, leading to substantial waste.

% 为此，人们想到可以将前馈网络原生地设计成稀疏激活结构，最有代表性的方式就是 Mixture-of-Experts。MoE 将前馈网络沿着 intermediate dimension 拆分成多个子网络，并在词元进入这些子网络之前进行路由，让词元只会进入到有限个子网络中。这种设计既保证了模型具有很大规模的总参数，又让每个词元的前向传播只使用很小的参数量，兼顾了规模和效率。

To address this, researchers have proposed designing feed-forward networks with a natively sparse-activation structure, most notably the Mixture-of-Experts (MoE)~\citep{shazeer2017,switchtransformers}. MoE splits the feed-forward network into multiple sub-networks along the intermediate dimension and routes each token to only a limited number of them before computation. This design endows the model with a large total parameter count while ensuring that each token’s forward pass computation uses only a small fraction of parameters, reconciling scale with efficiency.

% 随着更加稀疏 MoE 的出现，模型在宽度维度上的参数效率正在被逐步提升。但在本文中，我们关心如何从深度维度提升模型参数效率。一方面，xxx 等一类可解释工作中指出，前馈网络中负责存储静态知识，并且发现不同层之间有可能存储相似的知识，存在一定的冗余。同时，xxx 等 Looped Transformer 相关的工作发现，这类架构可以使用显著更小的参数量达到更大参数量的推理效果。不过 Looped Transformer 为此使用了更多次的递归和更高的串行度，付出了显著更高的计算和时间开销作为代价。

With the advent of increasingly sparse MoE models, parameter efficiency along the width dimension has been steadily improved~\citep{dai2024deepseekmoe}. In this work, however, we ask how parameter efficiency can be improved along the depth dimension~\citep{raposo2024mixture}. On the one hand, interpretability studies suggest that feed-forward networks store relatively static knowledge~\citep{geva-etal-2021-transformer,dai-etal-2022-knowledge}, and that different layers may store similar knowledge, suggesting nontrivial redundancy across layers~\citep{rome,takase2023lessons,fan2019reducing,chen2025streamlining,men2025shortgpt}. On the other hand,
Looped Transformers reuse the same parameters across recurrent applications
of a block~\citep{dehghani2018universal,giannou2023looped,bai2019deep,csordas2024moeut,chen2026intern},and recent work
shows that this design can attain the reasoning performance of much larger
models with far fewer parameters~\citep{bae2025relaxed,
saunshi2025reasoning,wang2026smelt}. This benefit, however, comes with
additional recursions and greater serial depth, incurring higher computational
and latency costs.

% On the other hand, xxx and related works on Looped Transformers have found that such architectures can achieve the reasoning performance of much larger models with far fewer parameters; however, Looped Transformers pay a substantial cost in additional recursions and higher serial depth, incurring markedly higher computational and time overhead.

% 为了兼顾参数存储效率和计算开销，本文尝试探索了另一种 Parameter Loop 方案的潜力 —— Cross-Layer Expert Sharing。与传统的 Looped Transformer 维持每层的总参不变、但增加递归的次数不同，我们希望保持递归的次数不变，但是提升每层的候选专家数，并通过 MoE 的稀疏激活来保持每层的激活参数和计算量不变。我们称这部分层间共享的专家为 Meta Experts，称传统设计中层内私有的专家为 Local Experts。

To balance parameter storage efficiency and computational cost, this work explores the potential of another Parameter Loop scheme — Cross-Layer Expert Sharing. Unlike Looped Transformers, which keep the total parameters of each layer unchanged but increase the number of recursions, we aim to keep the number of recursions unchanged while increasing the candidate experts of each layer, and use the sparse activation of MoE to keep both the activated parameters and the computation per layer unchanged. We refer to the experts shared across layers as \textbf{Meta Experts}, and the layer-private experts in the conventional design as \textbf{Local Experts}. 

% 我们的实验发现，相比于传统的分层架构的 Local Experts，Meta Experts 可以为模型带来更低的参数冗余，并有着更低的交叉熵损失、更高的下游任务准确率。我们的进一步分析也解释了层间共享专家起效的另一种原因，Meta Experts 为 Hidden States 提供了更加灵活的存储读取路径。相比于传统的分层架构，层间共享专家隐式地提供了更灵活残差连接，它不仅有前向残差连接，还引入了现有模型都不具备的反向残差连接。

Our experiments show that, compared with the Local Expert of hierarchical architectures, Meta Experts reduce parameter redundancy and achieve lower cross-entropy loss together with higher downstream task accuracy. At a fixed number of activated experts and expert granularity, our model achieves comparable performance to Vanilla MoE with only 70\% as many experts, indicating a substantially higher effective capacity per parameter. 
Further analysis reveals another reason why cross-layer expert sharing is effective: Meta Experts provide hidden states with more flexible paths for storage and retrieval. Compared with hierarchical architectures, cross-layer expert sharing implicitly provides more flexible residual connections: beyond the forward residual connections, it introduces a backward residual connection at the activation level---one that operates within a single forward pass and is therefore absent from existing models.

% 综上，我们的贡献主要包括：1) 系统性分析了 Meta Experts 对于模型效果的影响。我们主要考虑了 Meta Experts 的激活量和总数量占所有专家比例、Meta Experts 是否在每层都存在、Meta Experts 是否也有一定的层级化等因素的影响。虽然我们发现，这些配置会适当影响模型的表现，但在绝大多数配置下，引入 Meta Experts 都对模型的能力有显著的提升; 2) 通过残差连接的视角给出了 Meta Experts 起效的一种解释。我们对比了是否包含 Meta Experts 的模型在前向传播时的 hidden states 更新机制和反向传播时的参数更新机制。我们的分析指出，现有的设计可以让模型在浅层和深层同时访问到浅层和深层的表征信息，带来了更加灵活的信息检索。

Our contributions are summarized as follows:
(1) A systematic analysis of the effect of Meta Experts on model
performance, covering their activation and total-number ratios among all
experts, their presence in every layer, and their partial hierarchical
organization. Although these configurations moderately affect performance,
Meta Experts consistently bring significant improvements.
(2) An explanation of why Meta Experts are effective from the perspective
of residual connections. By comparing the forward hidden-state updates and
the backward parameter updates of models with and without Meta Experts, we
show that this design grants shallow and deep layers bidirectional access
to representations across depth, enabling more flexible information
retrieval.

\section{Meta Expert}

% Local 和 Meta Experts 的实现方式。相比于每层私有的 Local Experts， Meta Experts 在所有层中共享。
\begin{figure}[ht]
    \centering
    \includegraphics[width=\textwidth]{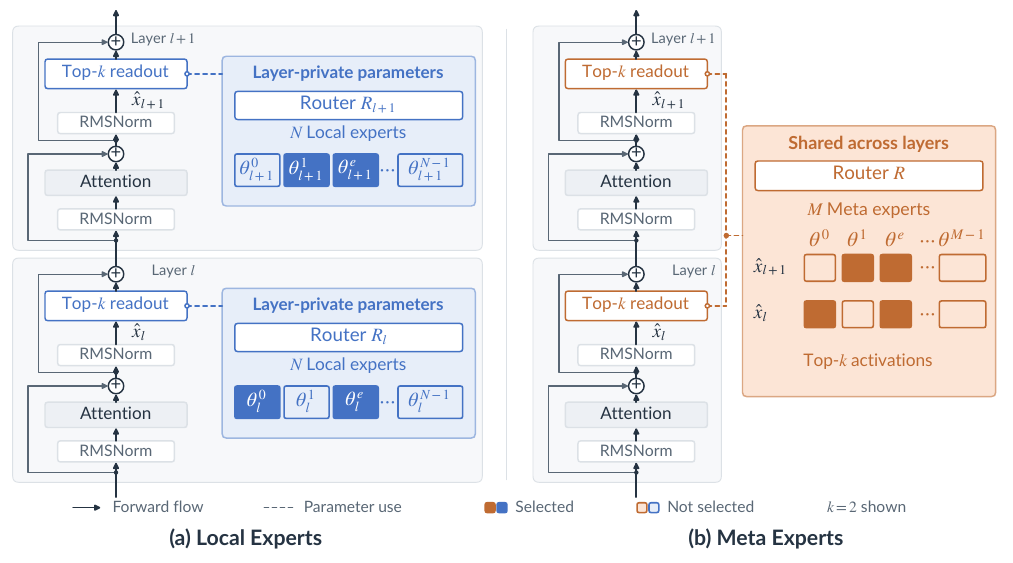} 
    \caption{The Implementation of Local and Meta Experts. In contrast to the layer-private
Local Experts, Meta Experts are shared across all layers.}
    \label{fig:main-arch}
\end{figure}

A MoE-based Transformer layer $l$ consists of an attention block and a MoE
block (Fig \ref{fig:main-arch}.(a)). The MoE block contains a layer-specific 
router $R_l$ and $N$ Local Experts $\{\phi_e(\cdot;\theta_l^e)\}_{e=0}^{N-1}$, 
both of which are initialized and accessed independently across layers. 
Given the hidden state $\hat{x}_l$ of a token entering the MoE block, the
router computes gating scores from its logits and selects the top-$k$
experts belonging to the current layer to process the token:

\begin{equation}
  y_l \;=\; \sum_{e \in \mathrm{topk}(R_l, \hat{x}_l)}
  R_l(\hat{x}_l)_e \, \phi_e(\hat{x}_l; \theta_l^e).
  \label{eq:local-moe}
\end{equation}

In a model built purely from Meta Experts (Fig \ref{fig:main-arch}.(b)), 
the layer-private MoE blocks are replaced with a single shared block: 
one router $R$ and one pool of $M$ experts
$\{\phi_e(\cdot;\theta^e)\}_{e=0}^{M-1}$ could be activated by every layer:

\begin{equation}
  y_l \;=\; \sum_{e \in \mathrm{topk}(R, \hat{x}_l)}
  R(\hat{x}_l)_e \, \phi_e(\hat{x}_l; \theta^e),
  \label{eq:meta-moe}
\end{equation}

\section{Empirical Comparison between Meta and Local Experts}

% 为了系统性分析 Meta Experts 对于模型表现的影响，我们以只包含 Local Experts 的模型作为起点，尝试以多种形式引入 Meta Experts，并观察引入前后的模型能力变化。

To systematically analyze the effect of Meta Experts on model performance,
we start from a model with only Local Experts and incorporate Meta Experts
in various forms, comparing model capabilities with and without them.

\subsection{Setups}

% 具体来讲，我们考虑了：1）Meta Experts 和 Local Experts 的数量关系，包括总数量比例和层内的激活比例；2）Meta Experts 和 Local Experts 的拓扑关系，比如，是否每层都包括 Meta Experts 和 Local Expert，以及 Meta Experts 是否也进行一定的层级化。

\begin{figure}[t]
    \centering
    \includegraphics[width=\textwidth]{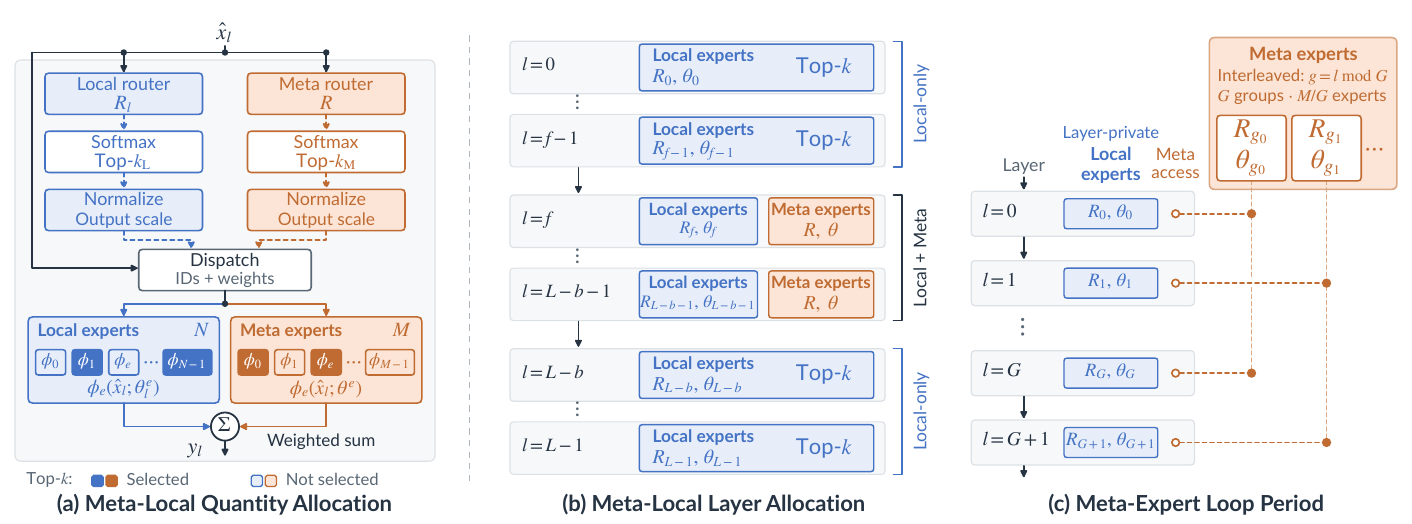} 
    \caption{The Quantitative and Topological Comparison between Meta and Local Experts.}
    \label{fig:meta-local-allocation}
\end{figure}

\textbf{Core Variables.}
We consider two types of relationship between Meta Experts and Local Experts: \textbf{(1) the quantitative relationship}: the ratio of their total numbers within the whole model and their activated numbers within each layer (See Fig \ref{fig:meta-local-allocation}.(a)); \textbf{(2) the topological relationship}: whether every layer contains both Meta and Local Experts (See Fig \ref{fig:meta-local-allocation}.(b)), and whether the Meta Experts are also hierarchically organized (See Fig \ref{fig:meta-local-allocation}.(c)). 

% 在实验过程中，我们先在一个 2B-A0.5B 的模型上遍历所有变量，然后再选择相对好的配置进行 Scaling Ladder 实验。在所有实验中，我们保持专家总数量和总激活量不变，这间接保证了不同模型具有相近的 FLOPs，从而进行更公平的对比。在对比不同 FLOPs 的模型效果时，我们让不同模型都使用 Vanilla MoE 结构下的 Training-Optimal 配置，而没有为我们的设计使用最优配置。

\textbf{Common Settings.}
During the experiments, we first sweep all the variables on a 2B-A0.5B
model, and then select the relatively better-performing configurations for
the scaling ladder experiments.
Throughout the experiments, we keep the total number of experts and the
total activation count fixed, which indirectly ensures that different
models have similar FLOPs, enabling a fairer comparison. When comparing
models at different FLOPs, all models adopt the training-optimal
configuration of the Vanilla MoE structure, rather than a configuration
tuned specifically for our design. Detailed experimental setups can be found in \ref{app:ladder-settings}

% 除了比较交叉熵损失和下游任务准确率之外，我们还设置了一个 Metric 来度量不同设置下，模型内部所有 Experts 之间的冗余程度变化。训练的具体配置细节和冗余度的度量方式详见附录。

% \textbf{Metrics.}
% In addition to comparing cross-entropy
% loss and downstream task accuracy, we design a metric to measure how the
% degree of redundancy among all experts inside the model varies under
% different settings. The exact training configurations and the definition of
% the redundancy metric are deferred to the appendix \ref{appendix:redundancy}.

\textbf{Metrics.}
In addition to cross-entropy loss and downstream task accuracy, we introduce
an expert-redundancy metric as a complementary perspective for analyzing
expert specialization under different architectural settings. It allows us to
examine how expert topology affects the overlap among learned expert
functions. The exact training configurations and the definition of the
redundancy metric are deferred to Appendix~\ref{appendix:redundancy}.

\subsection{Meta/Local Experts' ratio}
% 一个是 num_experts 的比例，一个是 num_activated_experts 的比例

% 我们首先比较了 Meta Experts 和 Local Experts 总数量比例和层内激活比例对于模型表现的影响。

We first study the effect of the ratio between the total numbers of Meta
Experts and Local Experts, and of their activation ratio within each layer,
on model performance. 

% 关于总数量比例，我们考虑了 Meta Experts 占比 0%/25%/50%/75%/100% 这几种情况；关于激活比例，我们考虑了 Meta Experts 占比 25%/50%/75% 这几种情况。加入给定一个 2 层的模型，总共有 256 个 experts，每层激活 8 个。那当 Meta/Local Experts 的数量占比 75% 时，Meta Experts 应该有 192 个，而每层 Local Experts 应该有 32 个。当 Meta/Local Experts 的激活占比 75% 时，每层激活 6 个 Meta Experts，激活 2 个 Local Experts。

\textbf{Settings.} 
For the total-number ratio, we consider Meta Expert proportions of
0\%, 25\%, 50\%, 75\%, or 100\%; for the activation ratio, we consider proportions
of 25\%, 50\%, or 75\%. For concreteness, consider a two-layer model with 256
experts in total and 8 experts activated per layer. A Meta/Local number
ratio of 75\% then corresponds to 192 Meta Experts, shared across all
layers, and 32 Local Experts per layer; an activation ratio of 75\%
corresponds to activating 6 Meta Experts and 2 Local Experts per layer.

\begin{figure}[ht]
    \centering
    \includegraphics[width=\textwidth]{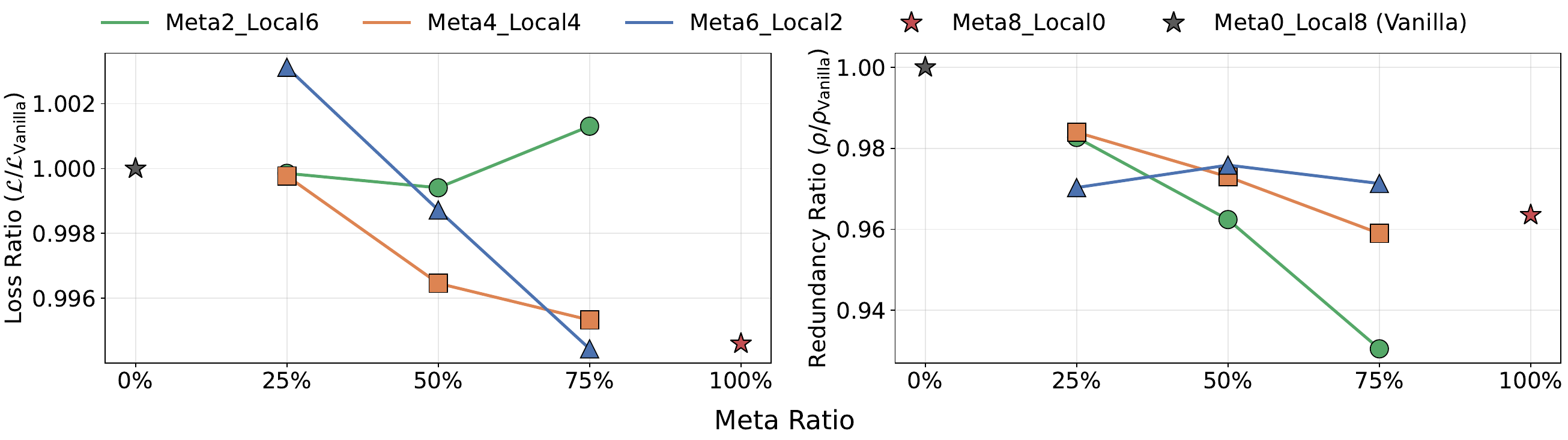} 
    \caption{Pilot experiments on the Meta/Local Expert ratio. We vary the
    total-number proportion of Meta Experts ($0\%$--$100\%$) and their
    per-layer activation proportion ($25\%$--$75\%$), while keeping the total
    expert count and the total activation count fixed so that all models share
    similar FLOPs. Higher Meta-Expert proportions consistently reduce both the
    cross-entropy loss and the expert redundancy $\rho$; increasing the
    activation share of Local Experts reduces redundancy but degrades the
    loss.}
    \label{fig:pilot-quantity}
\end{figure}

% 结果显示，在任何激活比例下，Meta Experts 的总参数占比越高，模型也会取得更低的 Loss 和 Redundancy，同时，为 Meta Experts 分配更高的激活比例也对 Loss 的降低有帮助，这些综合说明了 Meta Experts 的有效性。同时，虽然为 Local Experts 分配更高的激活比例对 Loss 是有损的，但对于降低专家间的 Redundancy 是有帮助的，这暗示着相比于 Meta Expert，传统架构中只在某一层被激活的 Local Experts 可能更难训练充分，并因而带来了 Redundancy。

\textbf{Pilot Experiments} (Fig \ref{fig:pilot-quantity}). 
The results show that, at any activation ratio, a higher total-parameter
proportion of Meta Experts consistently leads to lower loss and redundancy;
meanwhile, allocating a higher activation ratio to Meta Experts also helps
reduce the loss. Taken together, these observations attest to the
effectiveness of Meta Experts. On the other hand, while assigning a higher
activation ratio to Local Experts is detrimental to the loss, it does help
reduce the redundancy among experts. This suggests that, compared with Meta
Experts, Local Experts---each of which is activated in only one layer under
the traditional architecture---may be harder to train sufficiently, and
this under-training may give rise to redundancy.

\begin{figure}[ht]
    \centering
    \includegraphics[width=\textwidth]{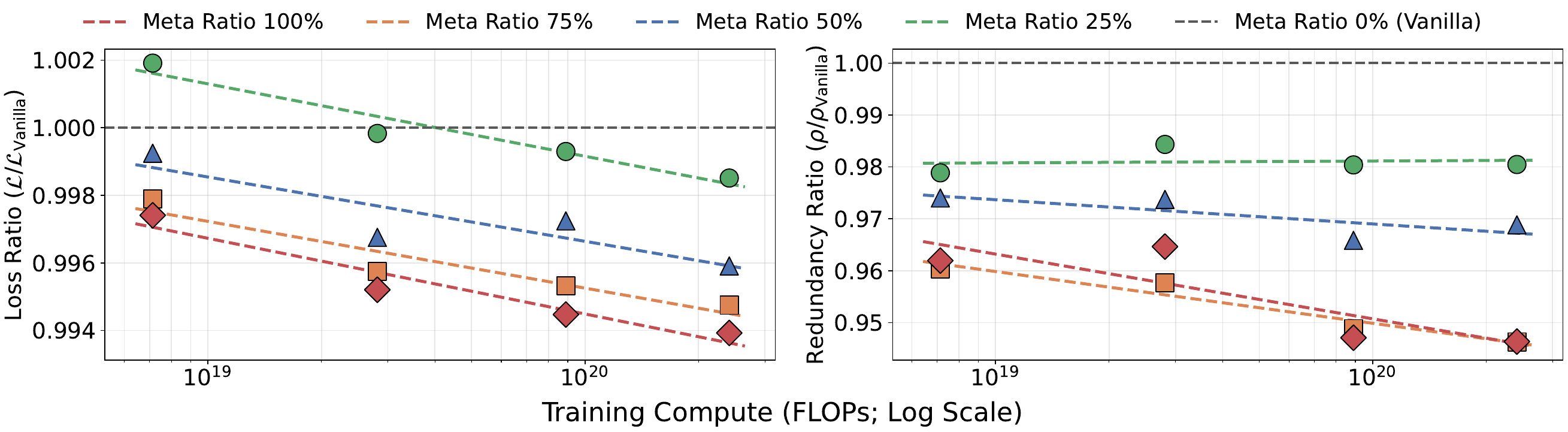} 
    \caption{Scaling-ladder on the Meta/Local Expert ratio. With the
    per-layer activation fixed to 4 Meta Experts and 4 Local Experts, we scale
    the total-number ratio of Meta Experts ($25\%$/$50\%$/$75\%$) across
    training compute (FLOPs). Although slightly worse than Pure-Meta-Expert Setting, 
    the $75\%$ ratio achieves the best cross-entropy loss over all Partial-Meta Experts
    setting at every compute budget, and is adopted as the base configuration in all subsequent experiments.}
    \label{fig:ladder-quantity}
\end{figure}

% 考虑到 Meta/Local Expert 都激活 4 的时候可以取得稳定更优的 Loss，我们在保持这个激活比例的情况下，对 25%/50%/75% 的总数量比例同步进行了完整的 Scaling Ladder 验证。经过验证可以发现，在各个 FLOPs 配置下，75% 都取得了最优效果，不管是交叉熵损失还是专家冗余度上。考虑到 75% Meta Experts 数量比例的优势，我们在后面的实验中都以此为基本配置。

\textbf{Scaling Ladder} (Fig \ref{fig:ladder-quantity}).
Given that activating 4 Meta Experts and 4 Local Experts per layer yields
consistently better loss, we hold this activation ratio fixed and conduct a
full scaling-ladder validation across the total-number ratios of
25\%/50\%/75\%. The results confirm that, although still slightly worse
than the Pure-Meta-Expert setting, the $75\%$ ratio achieves the best
performance over all Partial-Meta settings at every compute budget, in
terms of both cross-entropy loss and expert redundancy. Considering that 
a model with only Meta Experts may suffer from training instability, 
we adopt the 75\% configuration in all subsequent experiments.

\subsection{Meta/Local Experts' topology}

% 随后，我们探究了 Meta/Local Experts' topology 对模型表现的影响。我们一方面探索了是否每层都应该包括 Meta Experts，另一方面探索了 Meta Experts 是否也进行一定的层级化。

We next investigate the effect of the topology of Meta/Local Experts on
model performance. On the one hand, we explore whether every layer should
include Meta Experts; on the other hand, we explore whether the Meta
Experts should themselves be partially hierarchically organized.

% 一方面，我们考虑了在模型的仅前 1, 2, 4，8层，或者仅后 1, 2, 4，8 层，或者前后的各 1，2，4 层使用 Local Experts、完全不引入 Meta Experts 的情况。另一方面，我们考虑了将 Meta Experts 划分成 2, 4, 8 个子池，并考虑让这些子池按照 chunk 或者 interleave 的方式被 Looped 访问。

\textbf{Settings.}
On the one hand, we consider variants in which Local Experts are used
exclusively, with no Meta Experts introduced at all: either in only the
first 1, 2, 4, or 8 layers of the model, in only the last 1, 2, 4, or 8
layers, or in both the first and the last 1, 2, or 4 layers. On the other
hand, we consider partitioning the Meta Experts into 2, 4, or 8 sub-pools,
which are then accessed in a looped manner---either in a chunked or an
interleaved fashion.

\begin{figure}[ht]
    \centering
    \includegraphics[width=\textwidth]{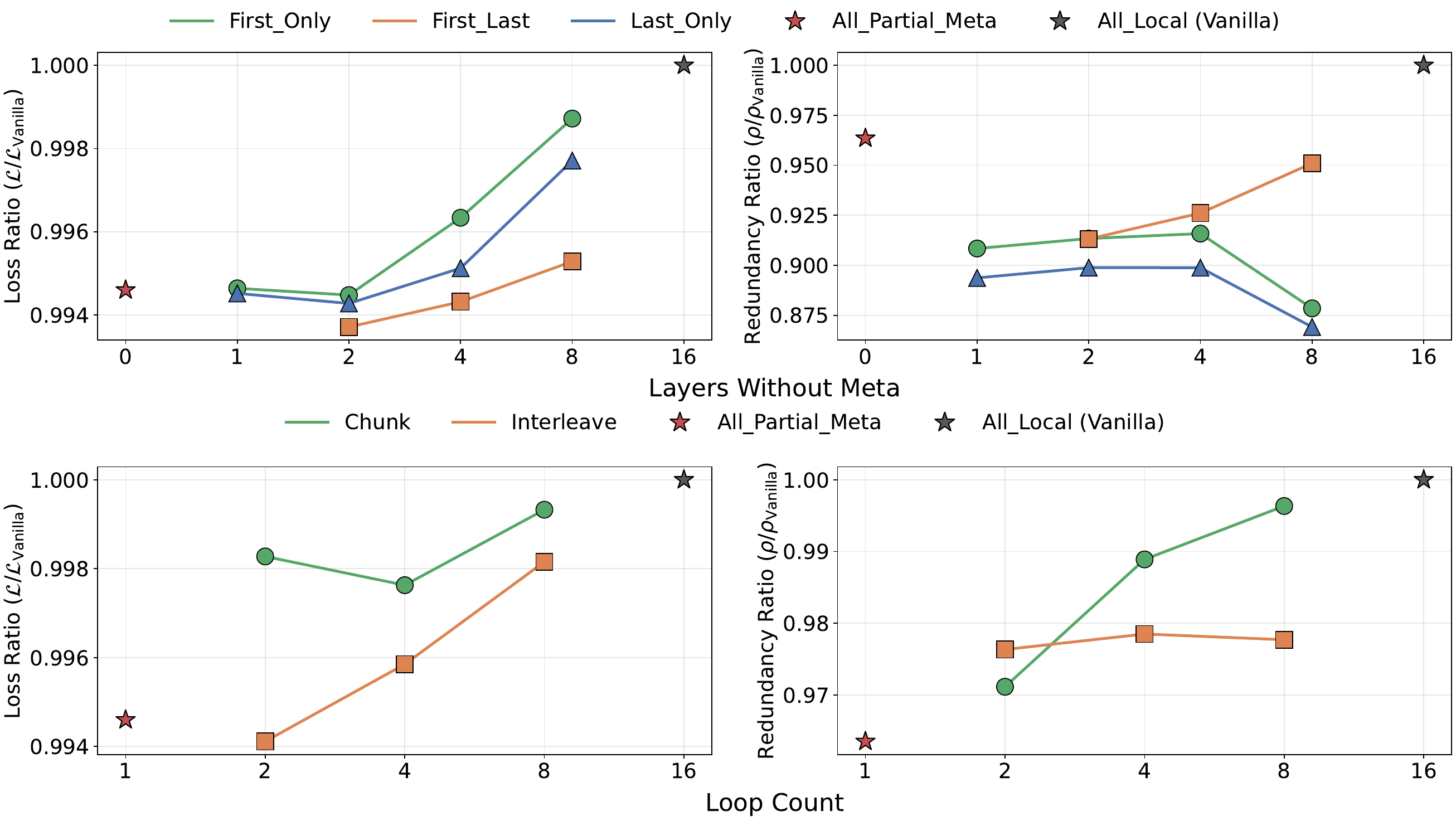} 
    \caption{Pilot experiments on the Meta/Local Expert topology. We vary
    where Local Experts are used exclusively (the first and/or the last
    1/2/4/8 layers), and how the Meta Experts are hierarchically organized
    ($2$/$4$/$8$ sub-pools, accessed in a looped manner, either chunked or
    interleaved). Both interventions generally degrade the loss, corroborating
    the benefit of Meta Experts, while a lighter degree of intervention---only
    the first and the last two layers restricted to Local Experts, or the Meta
    Experts looped only twice---slightly improves it.}
    \label{fig:pilot-topology}
\end{figure}

% 可以观察到这两类操作 —— 引入只包含 Local Experts 的层，以及模仿 Local Experts 的拓扑、将 Meta Experts 进行层级化 —— 普遍会带来 Loss 上的损失，这进一步说明了 Meta Experts 对模型表现的正面作用。但值得注意的是，在较少量的干预下，模型的 Loss 出现了轻微的降低，比如，只在靠前的 2 层和靠后的 2 层引入只包含 Local Experts，或者仅让 Meta Experts Loop 两次。这暗示我们，一方面，模型最浅层或最深层中的 Experts 有着差异化更为明显的功能，这些层之间基本不存在互相冗余的专家（通过观察专家冗余度的结果也可以获得类似的结论）。另一方面，人为地让 Meta Experts 具备轻度层级化有助于轻微地提升模型能力，但层级化会带来一定的专家冗余，但为什么这些冗余没有让模型的损失更高，有待进一步的探究。

\textbf{Pilot Experiments} (Fig \ref{fig:pilot-topology}).
It can be observed that both operations---introducing Local-Experts-only
layers, and organizing the Meta Experts hierarchically so as to mimic the
topology of Local Experts---generally incur a loss degradation, which
further demonstrates the positive role of Meta Experts in model
performance. Notably, under a lighter degree of intervention, the loss
slightly improves: for example, when only the first two and the last two
layers are restricted to Local Experts, or when the Meta Experts are looped
only twice. This suggests, on the one hand, that experts in the shallowest
and deepest layers serve more differentiated functions, with essentially
no mutual redundancy between these layers---a conclusion also supported by
the expert redundancy measurements. On the other hand, mildly hierarchical
Meta Experts help improve the model capability slightly; hierarchy,
however, introduces a certain amount of expert redundancy, and why this
redundancy does not translate into a higher loss remains an open question
for further investigation.

\begin{figure}[ht]
    \centering
    \includegraphics[width=\textwidth]{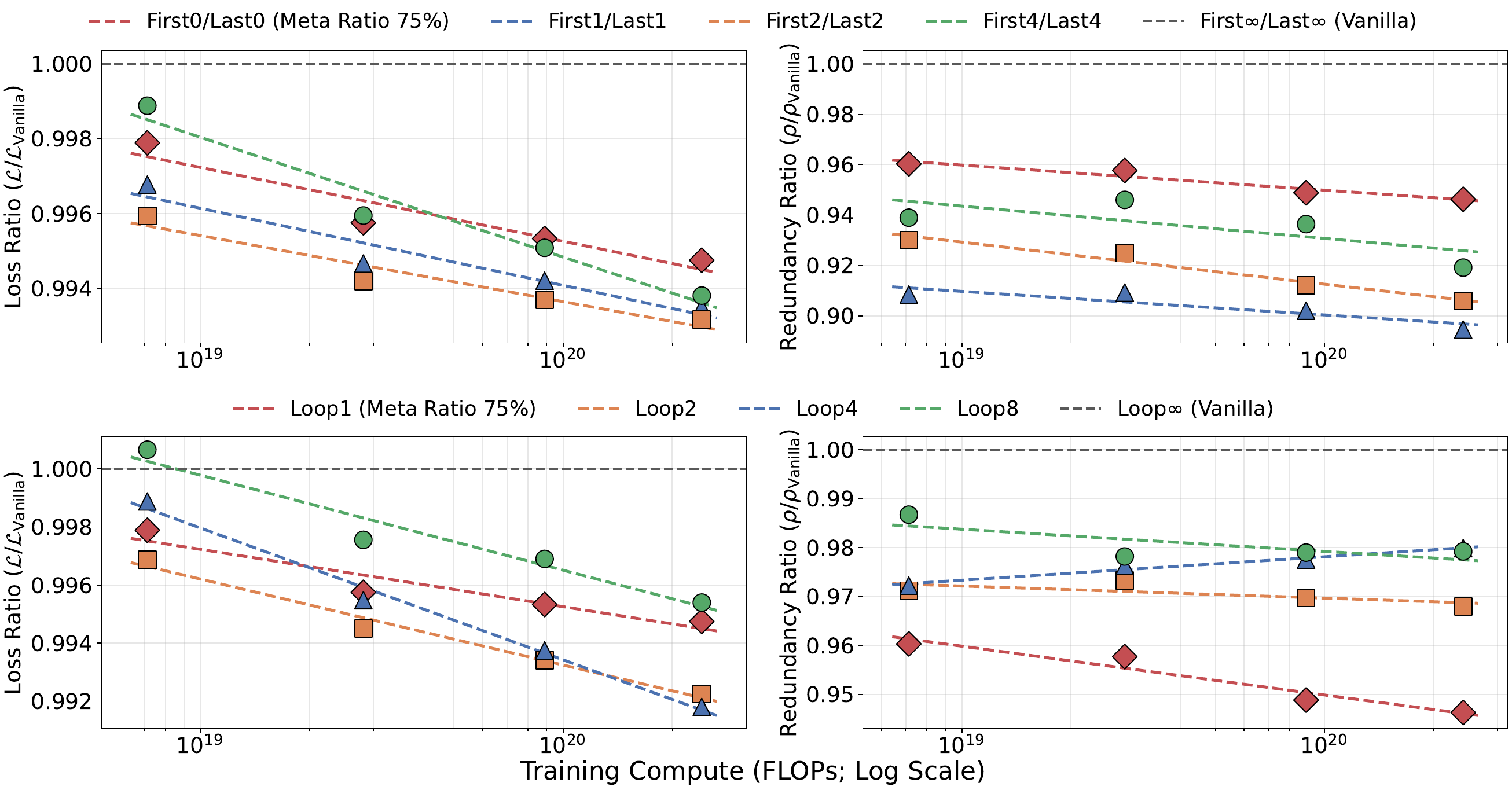} 
    \caption{Scaling-ladder on the Meta/Local Expert topology. We scale the two
    best pilot configurations---Local Experts exclusively at both shallow and
    deep ends, and interleaved sub-pool partitioning of the Meta
    Experts---across training compute (FLOPs). Restricting the first two and
    the last two layers to Local Experts, together with $2$ sub-pools,
    achieves the best performance at every compute budget; as the model
    becomes deeper, more Local-Experts-only layers and a larger hierarchy of
    Meta Experts further help, suggesting that deeper models call for a larger
    hierarchical expert topology.}
    \label{fig:ladder-topology}
\end{figure}

% 相较其他配置方式，同时在浅层和深层引入只包含 Local Experts 的层，以及使用 Interleave 的方式划分 Meta Experts 的子池，可以带来稳定更优的结果，我们在这两个配置下补充了完整的 Scaling Ladder 验证。结果表明，在各个 FLOPs 配置下，最优的效果由以下配置分别取得：前两层和后两层仅使用 Local Experts，以及将 Meta Experts 划分为 2 个子池。同时我们也观察到，随着模型层数的增加，使用更多只包括 Local Experts 的层，以及对 Meta Experts 使用更大的层级化，也展现出了更好的效果。这一方面暗示我们，模型的特异化浅层和深层占总层数的比例可能不随模型层数的改变而改变，另一方面，更深层的模型也需要更大的层次化 Expert 拓扑，但 Vanilla Model 中的 Local Experts 层次化程度过高，所以仅能取得次优的结果。

\textbf{Scaling Ladder} (Fig \ref{fig:ladder-topology}).
Compared with other configurations, introducing Local-Experts-only layers at
both the shallow and the deep ends, and partitioning the Meta Experts into
sub-pools in an interleaved manner, yield consistently better results; we
therefore supplement full scaling-ladder validation under these two
configurations. The results show that, at every FLOPs configuration, the
best performance is achieved by using Local Experts exclusively in the
first two and the last two layers, together with partitioning the Meta
Experts into 2 sub-pools. We also observe that, as the model becomes
deeper, using more Local-Experts-only layers and a larger hierarchy of
Meta Experts likewise exhibits better results. This suggests, on the one
hand, that the proportion of specialized shallow and deep layers relative
to the total number of layers may remain roughly constant as model depth
changes; on the other hand, deeper models also call for a larger hierarchical
expert topology; the Local Experts of the Vanilla model, however, are
overly hierarchical, and can therefore only achieve suboptimal results.

% 我们将各个消融下的最优方案进行了组合，发现组合后的模型在 Loss 上稳定由于各个配置。同时，以这个最优配置为基准，我们尝试减少 Meta Expert 的数量，发现相比 Vanilla 模型，具备 Meta Expert 的模型可以节省约 70-80% 的专家数（或 75-85% 的参数，因为其他模块也会占一定参数），并且随着模型参数量的变大，优势会变得更明显

\begin{figure}[ht]
    \centering
    \includegraphics[width=\textwidth]{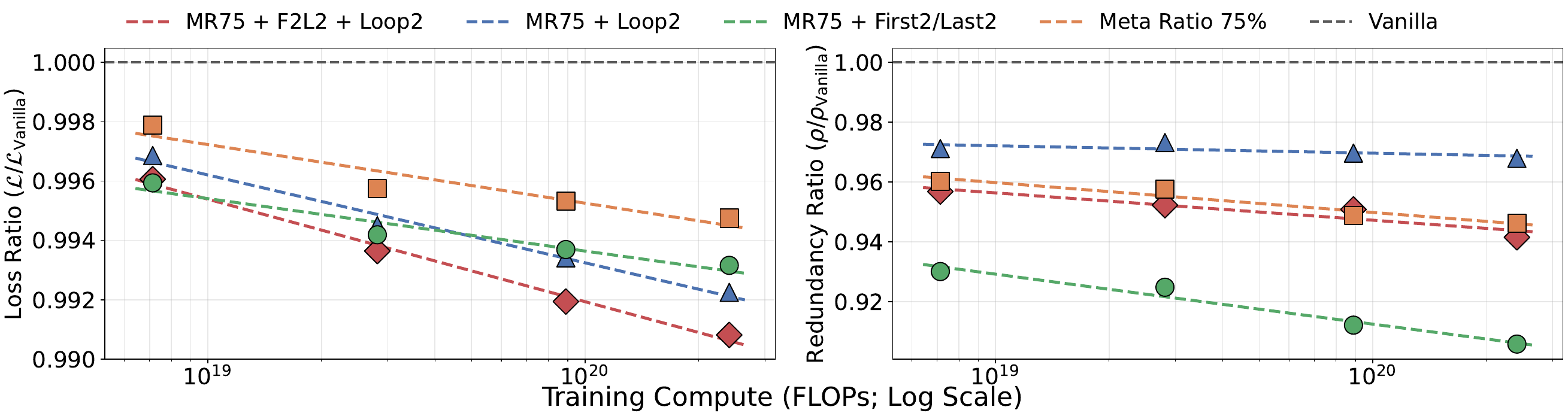} 
    \caption{Scaling-ladder on cumulative Meta-Expert designs. Starting from a
    $75\%$ Meta-Expert total-number ratio (MR75), we incrementally
    add topology refinements---Local Experts only in the first and last two
    layers (F2L2) and two looped Meta-Expert sub-pools (Loop2)---and
    finally combine both (F2L2 + Loop2). All models are compared against the
    Vanilla MoE (dashed line at $1.0$) in loss ratio $L/L_{\mathrm{Vanilla}}$
    (left) and expert redundancy ratio $R/R_{\mathrm{Vanilla}}$ (right) across
    training compute (FLOPs). Every refinement improves over Vanilla at every
    compute budget, with the largest gains under the combined design; the
    advantage in both loss and redundancy becomes more pronounced as training
    compute grows.}
    \label{fig:ladder-overall}
\end{figure}

\textbf{Overall Comparison} (Fig \ref{fig:ladder-paritial-experts} \& \ref{fig:ladder-overall})
We combine the best configurations identified in each ablation and find
that the combined model consistently outperforms every individual
configuration in terms of loss. Building on this best configuration, we
further reduce the number of Meta Experts and find that, compared with the
Vanilla model, models with Meta Experts save approximately 20--30\% of the
experts (or 15--25\% of the parameters, since other modules also account
for a certain fraction of the parameters), with the advantage becoming
more pronounced as the model size grows.

\subsection{Downstream Performance of Meta Experts-based Model}
% 为了更加全面的对比，在交叉熵损失之外，我们还尝试通过更大规模的训练来比较不同配置下，模型的下游任务能力。

For a more comprehensive comparison, beyond cross-entropy loss, we further
compare the downstream task capabilities of different configurations
through training at a larger scale.

% 我们使用了上面每个消融实验中的最优配置，并使用 6B-A1.6B 的模型训练了大约 400B tokens，用来观察他们在下游任务中的表现

\textbf{Settings.}
Using the best configuration identified in each of the ablation studies above, we train 6B-A1.6B models for 200B tokens and evaluate their downstream task performance. We compare five variants: the \emph{Vanilla} baseline (Vanilla MoE, Local Experts only), the \emph{MR75} model, which activates Meta Experts at a 75\% meta ratio, and three topological refinements of MR75---\emph{MR75\,+\,Loop2}, which partitions the Meta Experts into two interleaved sub-pools; \emph{MR75\,+\,F2L2}, which restricts the first two and last two layers to Local Experts only; and their combination, \emph{MR75\,+\,F2L2\,+\,Loop2}. All variants use the same total number of experts, per-layer activation budget, and training recipe, ensuring that performance differences are attributable to the expert topology. We evaluate
the models across a diverse set of benchmarks covering complementary
capabilities: world knowledge and factual question answering in both English
(MMLU~\citep{hendrycks2021measuringmassivemultitasklanguage}) and Chinese
(C-Eval~\citep{huang2023cevalmultilevelmultidisciplinechinese} and
CMMLU~\citep{li2024cmmlumeasuringmassivemultitask}); multi-step and commonsense
reasoning (BBH~\citep{kazemi2025bigbenchextrahard} with chain-of-thought and
OpenBookQA~\citep{mihaylov2018suitarmorconductelectricity}); and open-ended,
knowledge-intensive generation
(TriviaQA~\citep{joshi2017triviaqalargescaledistantly}).

\textbf{Results.}
Table~\ref{tab:downstream_qwen35_10k} reports the downstream performance of
the five expert-topology variants after training on 200B tokens. Due to the limited training budget, different Meta Expert variants exhibit advantages across different datasets, highlighting the need for larger-scale scaling experiments, but incorporating Meta Experts consistently outperform the vanilla baseline.

\begin{table}[t]
\centering
\small
\setlength{\tabcolsep}{5pt}
\renewcommand{\arraystretch}{1.2}
\caption{Downstream results for the 6B-A1.6B models after training on 200B
tokens. MMLU / C-Eval / CMMLU are 5-shot accuracy, OpenBookQA is normalized
accuracy, and BBH / TriviaQA are generation exact-match.
The best result per dataset is in \textbf{bold}. \emph{Vanilla} is the
Local-Experts-only baseline (Vanilla MoE); \emph{MR75} activates Meta Experts
at a 75\% meta ratio; \emph{+\,Loop2} partitions the Meta Experts into two
interleaved sub-pools; and \emph{+\,F2L2} restricts the first two and last two
layers to Local Experts only.}
\label{tab:downstream_qwen35_10k}
\begin{tabular}{lccccc}
\toprule
Dataset & Vanilla & MR75 & MR75 & MR75 & MR75 \\
        &         &      & +\,Loop2 & +\,F2L2 & +\,F2L2\,+\,Loop2 \\
\midrule
MMLU       & 0.4087 & \underline{0.4120} & \textbf{0.4175} & 0.3928 & 0.4052 \\
C-Eval     & 0.4354 & \underline{0.4443} & \textbf{0.4487} & 0.4160 & 0.4302 \\
CMMLU      & 0.4203 & 0.4161 & \textbf{0.4344} & 0.4198 & \underline{0.4249} \\
TriviaQA   & 0.1924 & 0.1964 & 0.1932 & \underline{0.2014} & \textbf{0.2042} \\
BBH        & 0.2972 & 0.2912 & \underline{0.3030} & \textbf{0.3061} & 0.2984 \\
OpenBookQA & 0.3500 & 0.3380 & \textbf{0.3640} & 0.3320 & \underline{0.3540} \\
% \midrule
% Average    & 0.3515 & 0.3511 & \textbf{0.3588} & 0.3447 & 0.3522 \\
\bottomrule
\end{tabular}
\end{table}

\section{Meta Experts as Bidirectional Residual Pathways}
\label{sec:residual-view}

Consider a stack of $L$ MoE layers with hidden size $H$. Writing $x_l$ for
the residual stream entering the MoE sub-layer of layer $l$,
$\hat{x}_l = \mathrm{RMSNorm}(x_l)$, and $\gamma$ for the residual scaling
coefficient, the sub-layer updates the stream as
\begin{equation}
  x_{l+1} \;=\; x_l \;+\; \gamma\, E_{\theta_l}(\hat{x}_l;\,R_l),
  \label{eq:layer}
\end{equation}
where $E_{\theta_l}$ is an expert block of $N$ SwiGLU experts, each with
parameter dimension $d$, $R_l$ is the router, and top-$k$ routing activates
$k$ experts per token. In a layer-wise MoE, $\theta_l$ and $R_l$ are
layer-private; in Meta Experts, a single block $\theta$ and router $R$ are
learned once and shared by all layers.

\subsection{Layer-wise MoE.}
\textbf{Forward (read).} The expert of layer $l$ injects its output into the
residual stream at layer $l$ (Eq.~\eqref{eq:layer}) and propagates only
forward: hidden state $x_m$ can contain layer $l$'s content only if $l<m$,
and even then only as a frozen component, because the readout operator of
layer $m$ ($E_{\theta_m}, R_m$) differs from the write operator of layer
$l$ ($E_{\theta_l}, R_l$).

\textbf{Training (write).} The residual connection does run backward as a gradient
highway: the adjoint $\bar{x}_l = \partial \mathcal{L} / \partial x_l$
inherits $\bar{x}_{l+1}$ through the identity branch with unit gain, so
the training signal of deeper layers does reach shallower parameters
$\theta_l$ without attenuation. This channel, however, is parameter-level
rather than activation-level, lands in layer-private parameters that only
layer $l$ ever reads, and over the $m-l$ hops between a pair of layers
$l<m$, the deep-layer component is progressively mixed into the composite
adjoint, diluted by the training signals of all intermediate layers rather
than arriving at $\theta_l$ as an isolated, addressable contribution.

\textbf{Net effect.} In one forward pass, a hidden state includes only
earlier layers' expert content; deeper layers' signal reaches a
shallow layer solely as a parameter update during training, never as
content in its readout.

\subsection{MoE with Meta Experts.}

\textbf{Training (write).}
Let $y_l = \gamma E_\theta(\hat{x}_l; R)$ and
$\bar{y}_l = \partial \mathcal{L} / \partial y_l$. Since all layers share
the same pool, layer $l$'s gradient contribution
\begin{equation}
  g_l
  \;=\;
  \Bigl(\frac{\partial y_l}{\partial \theta}\Bigr)^{\!\top}
  \bar{y}_l
  \;=\;
  \left.\frac{\partial \mathcal{L}}{\partial \theta}\right|_{l}
  \;\in\;
  \mathbb{R}^{Nd}
  \label{eq:gl}
\end{equation}
accumulates with all other layers' by the chain rule,
$\bar{\theta} = \sum_l g_l$ (the shared router obeys the same
accumulation; we focus on the expert parameters). Thus, with learning rate
$\eta$, after $T$ training steps the storage contains the superposition of
all layers' writes:
\begin{equation}
  \theta^{(T)}
  \;=\;
  \theta^{(0)}
  \;+\;
  \sum_{l=0}^{L-1} \bar{\Delta\theta}_l,
  \qquad
  \bar{\Delta\theta}_l
  \;=\;
  -\eta \sum_{t<T} g_l^{(t)}.
  \label{eq:theta-update}
\end{equation}
Equivalently, at training step $t$, layer $l$ contributes the local
parameter update
$\Delta\theta_l^{(t)} = -\eta g_l^{(t)}$.

\textbf{Forward (read).}
After training, every layer reads from the same shared expert parameters;
in particular, layer $m$ computes
$y_m = \gamma E_{\theta^{(T)}}(\hat{x}_m; R)$.
To characterize how a write from layer $l$ affects the readout at layer
$m$, we consider its local first-order effect at training step $t$,
evaluated at $\theta^{(t)}$ while holding the layer inputs fixed. Define
\begin{equation}
  J_m^{(t)}
  \;=\;
  \left.
  \frac{\partial E_\theta(\hat{x}_m^{(t)};R)}
       {\partial \theta}
  \right|_{\theta=\theta^{(t)}}
  \;\in\;
  \mathbb{R}^{H\times Nd}.
  \label{eq:read-jacobian}
\end{equation}
Since $y_l = \gamma E_\theta(\hat{x}_l;R)$,
Eq.~\eqref{eq:gl} gives
\[
  g_l^{(t)}
  \;=\;
  \gamma {J_l^{(t)}}^{\top}\bar{y}_l^{(t)}.
\]
Therefore, the write from layer $l$ at step $t$ induces the following
first-order change in layer $m$'s readout:
\begin{equation}
  \Delta y_{m\leftarrow l}^{(t)}
  \;\approx\;
  \gamma J_m^{(t)}\Delta\theta_l^{(t)}
  \;=\;
  -\eta\gamma^2
  J_m^{(t)}{J_l^{(t)}}^{\top}\bar{y}_l^{(t)}.
  \label{eq:read}
\end{equation}
Accumulating these local first-order influences over training gives
\begin{equation}
  \mathcal{I}_{m\leftarrow l}
  \;\coloneqq\;
  \sum_{t<T}\Delta y_{m\leftarrow l}^{(t)}
  \;\approx\;
  -\eta\gamma^2
  \sum_{t<T}
  J_m^{(t)}{J_l^{(t)}}^{\top}\bar{y}_l^{(t)}.
  \label{eq:kernel}
\end{equation}
The update of the shared router contributes an analogous term, omitted
because top-$k$ selection is piecewise constant in $R$.
The factor
$J_m^{(t)}{J_l^{(t)}}^{\top}$
has the form of a tangent kernel of the shared block evaluated
\emph{across} the two layers. The total shared-parameter influence on
layer $m$,
$\mathcal{I}_m = \sum_{l=0}^{L-1}\mathcal{I}_{m\leftarrow l}$,
contains contributions from \emph{every} layer, in contrast to both
limitations of layer-wise MoE:
\begin{itemize}
  \item \textbf{$l<m$ (deep reads shallow).} Versus the \emph{frozen}
        forward carrying of layer-wise MoE: because the write at layer $l$
        and the readout at layer $m$ are two evaluations of the
        \emph{same} operator, layer $m$ retrieves, re-encodes, and further
        transforms shallow content in the code space in which it was
        written.
  \item \textbf{$l>m$ (shallow reads deep).} Versus the \emph{gradient
        highway} of layer-wise MoE: deep content reaches layer $m$ not as a
        parameter update of its private $\theta_m$, but as an
        activation-level term in its readout, in the common code space,
        through an $O(1)$-length path independent of $m-l$.
\end{itemize}

\textbf{Net effect.}
A single forward pass provides bidirectional access along depth---deep
hidden states reach shallow expert content, and shallow hidden states
reach deep expert content---through one shared storage. More discussions
are done in the Appendix \ref{appendix:discussion-residual}.

\section{Conclusion}
\label{sec:conclusion}

In this work, we present an empirical and theoretical analysis of
cross-layer expert sharing in MoE models, contrasting the hierarchically
arranged \textbf{Local Experts} of standard architectures with inter-layer
shared \textbf{Meta Experts}. Our systematic comparisons---over the
Meta/Local ratio, the activation ratio, and the topology of the shared
pool---show that Meta Experts consistently reduce expert redundancy and
cross-entropy loss while improving downstream task accuracy: at a fixed
number of activated experts and expert granularity, models built on shared
experts match the performance of Vanilla MoE with only 70--80\% of the
expert budget, substantially improving scaling potential. Beyond these
empirical observations, our analysis shows that a shared expert block is
not merely a parameter-saving device: it widens the forward residual
connections and grants hidden states bidirectional access to expert content
across depth, improving both the storage and the retrieval of information.
Taken together, our results suggest that inter-layer redundancy, long
treated as an intrinsic cost of deep MoE models, can instead be converted
into effective capacity. Future directions include extending this analysis
to larger models and broader architectures, and understanding why mildly
hierarchical shared experts introduce redundancy without degrading
performance.

\subsection*{AI use statement}

In this work, we used generative AI tools for writing assistance:
polishing language and grammar, brainstorming alternative phrasings (e.g., for the
paper title and section openings), formalizing the presentation of the
theoretical analysis (e.g., \LaTeX{} derivations whose underlying ideas are
the authors' own), and drafting descriptive passages (figure captions, the
Related Work section, and the Conclusion) based on author-provided
outlines, notes, and experimental results. We also used generative AI
tools for code assistance, including implementation support and plotting
scripts for the result figures; all designs, algorithms, and experimental
ideas were proposed by the authors.

We have not used generative AI tools for research ideation, experimental
design. We have reviewed all AI-assisted work: all AI-drafted text was checked, 
revised, and approved by the authors; translations were verified against 
the authors' intended meaning; all technical claims, equations, and 
reported numbers were independently verified by the authors against their 
own derivations and experimental results; and AI-generated code was verified, 
tested, and debugged by the authors for correctness, with all result figures 
manually checked against the underlying data. We take responsibility for the 
final content of this work, including text, claims or artifacts produced with
the aid of generative AI.

% \subsection*{Ethics statement}

% (This section is \textbf{recommended} and does not count toward the page limit.)

% If authors feel that their paper submission raises questions regarding the Code
% of Ethics, they are encouraged to include a paragraph of Ethics Statement (at
% the end of the main text before references) to address potential concerns where
% appropriate. Topics include, but are not limited to, studies that involve human
% subjects, practices to data set releases, potentially harmful insights,
% methodologies and applications, potential conflicts of interest and sponsorship,
% discrimination/bias/fairness concerns, privacy and security issues, legal
% compliance, and research integrity issues (e.g., IRB, documentation, research
% ethics). This statement should not be more than 1 page.

\subsection*{Reproducibility statement}
We have taken several steps to make the results in this paper reproducible, and
we reference here the parts of the paper that support them. The complete
experimental configuration for the scaling ladder is given in
Appendix~\ref{app:ladder-settings}: the per-depth architecture is listed in
Table~\ref{tab:ladder-arch} and the per-depth training schedule---global batch
size, optimizer steps, warmup and decay, learning rate, token budget, and
training FLOPs---in Table~\ref{tab:ladder-compute}. Every rung, and every
architectural variant at a given rung, shares the same backbone, optimizer, and
parallelism policy; only the depth-dependent knobs in those tables change, so a
single set of hyperparameters fully specifies each run. The evaluation protocol
used for the compute-versus-loss analysis is described in the same appendix: for
each run we report the language-modeling loss averaged over the final ten
training steps. Our models are implemented on top of Megatron-LM, a widely used open-source framework, which helps ensure reproducibility.



\bibliography{iclr2027_conference}
\bibliographystyle{iclr2027_conference}

\appendix
\section{Appendix}

\subsection{Related Work}
\label{appendix:related}

\noindent\textbf{Weight Tying} was originally introduced to share
parameters between the embedding layer and the language model
head~\citep{press2017using,inan2016tying}: since the two matrices map tokens
to and from the hidden space, tying them nearly halves the corresponding
parameters while often improving perplexity, and the practice has become
standard in modern language models~\citep{vaswani2017attention,raffel2020exploring}.
Beyond tying modules at the same depth, weight tying can be pushed further,
to reusing parameters \emph{across} depth~\citep{dehghani2018universal,
lan2019albert}.

\noindent\textbf{Looped Transformer} pushes weight tying to extreme:
the same block is applied multiple times per token, unrolling a small set
of weights into arbitrary depth~\citep{dehghani2018universal,
giannou2023looped,hua2025intuitive}. This line of work shows that weight reuse
across depth can substitute for parameter growth, and that a compact
recurrent model can match the reasoning performance of much larger ones
\citep{bae2025relaxed,saunshi2025reasoning,wang2026smelt}, at
the cost of additional recursions and a higher degree of serialization.
Meta Experts share the same motivation---extracting more effective capacity
from a fixed parameter budget---but apply tying at a finer granularity and
within a single feed-forward pass: only the expert block is reused across
layers, without extra recursions, so training and inference retain the
parallelism of standard Transformers, while attention may remain
layer-specific. Depending on \emph{what} is reused across depth, this
paradigm subsumes several lines of work.

\noindent\textbf{Attn Weight Sharing} reuses the attention projections (or
entire attention blocks) of different layers, yielding moderate parameter
savings at little accuracy cost~\citep{xiao2019sharing,lan2019albert,
mu2026cross}. These results suggest that the per-layer
parameter budget of Transformers is partially redundant---but leave open
\emph{where} the redundancy concentrates and whether the same principle
applies to the rest of the block. \textbf{Attn KV-Cache Sharing} reuses not
parameters but activations: in long-context or serving scenarios, multiple
layers attend to a common KV cache, trading representational independence
for inference memory and throughput~\citep{sun2024you,
brandon2024reducing,zuhri2025mlkv,wu2024layer,
wu2025systematic}. Such sharing is typically applied at deployment
time, affects the cache rather than the learned weights, and is often found
to be nearly lossless---suggesting that layers can tolerate \emph{reading}
the same content, but not that they should \emph{store} the same content.
Meta Experts operate on the storage side: the shared expert pool is written
by every layer during training, so cross-layer sharing reshapes what the
model stores and how hidden states retrieve it, rather than merely reusing
cached activations at inference.

\noindent\textbf{FFN Weight Sharing} is the closest line to our work: it
reuses feed-forward parameters across layers, either by tying the FFN of
all layers or by sharing it among a subset~\citep{lan2019albert,
pires2023one,thawakar2024mobillama}. Existing
studies, however, focus on dense FFNs and report mixed results, and sharing
is typically applied uniformly to every layer. \textbf{MoE Weight Sharing} has been explored in
several recent works: experts are shared across layers, but almost always
under an every-layer sharing topology, whose communication and memory
overhead is difficult to scale~\citep{li2025megrez2,tan2025rexmoe,chen2026intern,
huang2026unipool,chen2026mixture,jiao2026csmoe}. Meta Experts
generalize this idea to a more flexible regime: a shared \emph{pool} of
experts is accessed through several topology that is more friendly for
communication and memory (reserving boundary layers for local experts,
sub-pool partitioning with chunked or interleaved access), allows the model
to decide which content is worth sharing across depth---and, as our analysis
shows, implicitly turns the shared experts into bidirectional residual
pathways between layers.

\subsection{Redundancy Metric}
\label{appendix:redundancy}

To quantify how much of a model's expert capacity is redundant, we need a
measure that is (i) \emph{continuous}, with no similarity threshold; (ii)
\emph{comparable} across architectures with different expert counts and
depths; and (iii) \emph{faithful} to the correlation structure among
experts in high dimensions, where naive cosine similarities collapse toward
zero. Our metric is built from a single similarity kernel and a single
spectral statistic.

\paragraph{Similarity kernel: linear-CKA on centered covariances.}
For expert $i$, let $Y_i \in \mathbb{R}^{T \times H}$ be its response
matrix to a shared probe set of $T$ L2-normalized tokens, and let $Y_i^c$
denote $Y_i$ with its sample mean subtracted. We map each expert to the
vectorized centered covariance
\begin{equation}
  \phi_i \;=\; \mathrm{vec}\bigl({Y_i^c}^{\top} Y_i^c\bigr),
  \label{eq:phi}
\end{equation}
and define the pairwise similarity as the cosine in this feature space,
\begin{equation}
  K_{ij} \;=\; \frac{\langle \phi_i, \phi_j \rangle}
                    {\|\phi_i\|\, \|\phi_j\|},
  \label{eq:cka}
\end{equation}
which is exactly linear CKA evaluated on the experts' functional responses.
Being invariant to orthogonal transformations and isotropic scaling,
$K_{ij}$ reflects the correlated directions between experts rather than
their absolute orientation or norm, and avoids the collapse of raw
high-dimensional cosine. The resulting Gram matrix
$K \in \mathbb{R}^{E \times E}$ covers the $E$ experts under analysis:
for models mixing Meta and Local Experts the two pools are concatenated on
the expert axis and analyzed jointly, and for layer-wise models the layer
index serves as the grouping axis.

\paragraph{Spectral statistic: how many directions collapse.}
Since $K$ is positive semi-definite, its eigen-spectrum
$\lambda_1 \ge \lambda_2 \ge \dots \ge \lambda_E \ge 0$ distributes the
total similarity energy across $E$ principal directions. We measure the
concentration of this energy by the Gini coefficient of the spectrum,
\begin{equation}
  \rho \;=\; \mathrm{Gini}(\lambda)
       \;=\; \frac{\sum_{i,j} |\lambda_i - \lambda_j|}
                  {2E \sum_i \lambda_i}
       \;\in\; [0, 1],
  \label{eq:gini}
\end{equation}
where $\rho = 0$ means the energy is uniform across all directions (experts
fully complementary, no redundancy) and $\rho \to 1$ means it concentrates
on a few directions (strong redundancy). The reading admits an exact
interpretation: for a uniform rank-$k$ spectrum,
$\mathrm{Gini}(\lambda) = 1 - k/E$, so $\rho$ equals the \emph{fraction of
expert directions that have collapsed} into redundant ones. Because $\rho$
is a ratio, it is directly comparable across models with different $E$,
depths, and training budgets, and we report it as our headline
\emph{Expert Redundancy} metric.

\subsection{Scaling-ladder experimental settings}
\label{app:ladder-settings}

\paragraph{Models and training.}
All models are trained from scratch on NVIDIA H200 GPUs. The scaling experiments
use a single family of Qwen3.5-style, text-only Mixture-of-Experts (MoE) models
arranged as a depth-controlled ladder at four depths $L\in\{12,16,20,24\}$; model
width is scaled jointly with depth so that each rung is (approximately)
compute-optimal, giving a clean compute-versus-loss axis. At each rung, the
vanilla baseline and every Meta Experts variant are trained independently,
sharing the same backbone architecture and training configuration but no trained
weights, and preserving the same $128\times L$ routed-expert budget, so only the
depth-dependent knobs (width, batch size, learning rate, token budget, and
expert parallelism) change across rungs (Tables~\ref{tab:ladder-arch}
and~\ref{tab:ladder-compute}).

\begin{table}[h]
  \centering
  \caption{Per-depth architecture of the scaling ladder. Every rung uses
    $128$ routed experts plus one gated shared expert, top-$8$ routing, a
    period-4 GDN/full-attention pattern (one quarter full attention), and a
    head dimension of $256$. Widths scale jointly with depth. The table
    specifies the shared vanilla backbone at each rung; the Meta Experts
    variants replace only the routed-expert layers and keep the same
    $128\times L$ routed-expert budget, so every variant at a given rung shares
    these dimensions. GDN key/value entries are head \emph{counts} (each head
    has dimension $128$).}
  \label{tab:ladder-arch}
  \begin{tabular}{lcccc}
    \toprule
    & \multicolumn{4}{c}{Depth $L$} \\
    \cmidrule(lr){2-5}
    Hyperparameter & 12 & 16 & 20 & 24 \\
    \midrule
    Hidden size $d_{\text{model}}$      & 768 & 1024 & 1280 & 1536 \\
    Attention heads                     & 6   & 8    & 10   & 12   \\
    GQA query groups                    & 1   & 1    & 1    & 2    \\
    Full-attention layers ($L/4$)       & 3   & 4    & 5    & 6    \\
    GDN key/value heads                 & 6/12 & 8/16 & 10/20 & 12/24 \\
    Routed experts                      & 128 & 128  & 128  & 128  \\
    Routed-expert FFN width             & 192 & 256  & 320  & 384  \\
    Shared-expert FFN width             & 192 & 256  & 320  & 384  \\
    Router top-$k$                      & 8   & 8    & 8    & 8    \\
    \bottomrule
  \end{tabular}
\end{table}

Each model is a hybrid-attention MoE transformer:
attention layers alternate in a fixed period-4 pattern of three gated-delta-net
(GDN) linear-attention layers followed by one full-attention layer, so one
quarter of the layers use full softmax attention---with grouped-query attention
and a head dimension of $256$---while the GDN layers use a key/value head
dimension of $128$ with a depthwise causal convolution of kernel width $4$. Every
transformer block carries an MoE feed-forward layer with $128$ routed experts and
one gated shared expert, whose intermediate widths scale with model width, and
the router selects the top-$8$ experts per token with pre-softmax gating.

We
optimize with Muon (extra scale factor $0.2$) under a warmup--stable--decay (WSD)
schedule with an inverse-square-root decay shape: the peak learning rate
decreases with depth (from $1.97\times10^{-3}$ at $L{=}12$ to
$1.278\times10^{-3}$ at $L{=}24$), the minimum is one tenth of the peak, warmup
spans tens of steps, and the decay phase spans the final $\sim\!10\%$ of
training; all models use a micro-batch size of $2$ and a sequence length of
$8192$. The reported
loss is the training language-modeling loss averaged over the final ten steps, so
all rungs are compared on the same quantity, and the per-rung token budget is
$\mathrm{GBS}\times8192\times\text{iters}$ (Table~\ref{tab:ladder-compute}).

All
rungs use tensor-, pipeline-, and context-parallel size $1$, scaling
by data and expert parallelism only: the two smaller rungs ($L\in\{12,16\}$) use
expert-parallel size $1$ with the all-gather token dispatcher, and the two larger
rungs ($L\in\{20,24\}$) use expert-parallel size $8$, the all-to-all dispatcher,
and block activation recomputation on the last $4$ layers, with grouped GEMM and
permute fusion enabled throughout. As an illustrative parameter scale, the
$L{=}16$ vanilla backbone has $\approx\!1.96$B total parameters and
$\approx\!449$M activated parameters per token.

\begin{table}[h]
  \centering
  \caption{Per-depth training schedule and compute of the scaling ladder. Batch
    size, learning rate, and token budget depend only on depth. GBS is the
    global batch size in sequences of length $8192$; the token budget is
    $\mathrm{GBS}\times 8192\times\text{iters}$. FLOPs are the training-compute
    values used for the compute-versus-loss analysis. Warmup and WSD-decay are
    given in optimizer steps.}
  \label{tab:ladder-compute}
  \begin{tabular}{lcccc}
    \toprule
    & \multicolumn{4}{c}{Depth $L$} \\
    \cmidrule(lr){2-5}
    Setting & 12 & 16 & 20 & 24 \\
    \midrule
    Global batch size (seqs)   & 96 & 160 & 256 & 384 \\
    Micro-batch size           & 2  & 2  & 2  & 2  \\
    Training steps             & 6747 & 8436 & 9763 & 11101 \\
    Warmup steps               & 43 & 26 & 16 & 11 \\
    WSD-decay steps            & 675 & 844 & 976 & 1110 \\
    Peak LR ($\times10^{-3}$)  & 1.970 & 1.664 & 1.444 & 1.278 \\
    Min LR ($\times10^{-4}$)   & 1.970 & 1.664 & 1.444 & 1.278 \\
    Token budget (B)           & 5.31 & 11.06 & 20.47 & 34.92 \\
    Training FLOPs & \(7.135\times10^{18}\) & \(2.813\times10^{19}\) & \(8.895\times10^{19}\) & \(2.413\times10^{20}\) \\
    \bottomrule
  \end{tabular}
\end{table}

\paragraph{Hyperparameter selection.}
% 不按架构变体调参，每档为 vanilla 基线固定一套计算最优配置并原样复用到各 partial-Meta 变体。该配置通过扫描深度-宽度缩放、全局批大小、峰值学习率和 token 预算得到，遵循计算最优训练与缩放律的既有实践，MoE 结构选择也参照近期 MoE 缩放律分析。由于变体不重新调参，损失差异反映架构变化而非超参数差异；且同档各变体路由和激活参数相同、token 预算相同，训练 FLOPs 相同，计算量-损失比较是等 FLOP 的。
We do not tune the training hyperparameters per architectural variant. Instead we
fix a single compute-optimal configuration for the vanilla baseline backbone at
each rung and reuse it unchanged for every Meta Experts variant. The baseline
configuration---the joint depth--width scaling, and the per-depth global batch
size, peak learning rate, and token budget---is obtained by sweeping these
settings and retaining the compute-optimal choice at each depth, following
established practice on compute-optimal training and its schedule
\citep{hoffmann2022training, hagele2024scaling} and on scaling laws for the
learning rate and batch size \citep{bi2024deepseek, team2025every}. The
Mixture-of-Experts structural choices (expert count, granularity, and the
shared-expert ratio) are likewise set in line with recent MoE scaling-law
analyses \citep{tian2026towards, zhao2025towards}. Because every
Meta Experts variant inherits the baseline configuration without re-tuning, a
difference in loss at a rung reflects the architectural change rather than a
difference in hyperparameters; this reuse is further supported by evidence that
optimal learning rate and batch size are governed primarily by the compute
budget and are largely insensitive to Mixture-of-Experts architectural
variation \citep{team2025every}. Moreover, at a given depth every variant keeps
the per-token top-$8$ routing and hence the same activated parameters, and shares
the same token budget, so all variants at a rung incur the same training FLOPs
and are plotted at a single shared compute coordinate; the compute-versus-loss
comparison at that rung is thus iso-FLOP by construction, and the only free axis
across variants is the loss.

\paragraph{Meta Expert configurations.}
Table~\ref{tab:ladder-variants} catalogues every architecture actually run in
the ladder---the vanilla baseline and each model incorporating Meta Experts---each swept over
all four depths $L\in\{12,16,20,24\}$. A Meta Expert variant changes
\emph{only} the routed-expert layers of the shared backbone
(Table~\ref{tab:ladder-arch}): it splits each layer's $128$ routed experts into a
\emph{Meta} pool that is shared across depth and a set of \emph{Local} experts
private to that layer, and routes each token to $m$ meta $+\,l$ local experts. We
write a split as (meta width\,/\,local width) per layer; the recurring
\emph{r75} base is $96$ meta $/\,32$ local with $m{+}l = 4{+}4$. Two invariants
hold for every variant at every depth and render the rungs comparable: the total
routed-expert count is $128\times L$ (iso-param, as the shared Meta pool is
counted once across the $L$ layers) and the per-token routing is top-$8$
(iso-FLOP). The attention stack, gated shared expert, and training configuration
are identical to the vanilla baseline, so every variant at a rung occupies the
single shared compute coordinate of Table~\ref{tab:ladder-compute}.

The variants span two axes. The \emph{quantitative} axis is the
\emph{ratio} family, which varies the meta share of each layer's $128$ experts
over $25/50/75\%$---MR25/MR50/MR75, i.e. $32/64/96$ meta with the complementary
$96/64/32$ local---while routing $4$ meta $+\,4$ local per token. The
\emph{topological} axis fixes the r75 split and varies where and how the Meta
pool is shared across depth. The \emph{first/last local-only} family converts the
first and last $1/2/4$ layers (\texttt{f1l1}/\texttt{f2l2}/\texttt{f4l4}) to
$128$ Local-only layers---no meta, with all $8$ routed among the locals---while
the interior layers retain r75; \texttt{f2l2} is the F2L2 point of the main text.
The \emph{meta-loop} family retains r75 at every layer but replaces the single
globally shared Meta pool with $2/4/8$ interleaved sub-pools, each reused by
every $2$nd$/4$th$/8$th layer (Loop2/4/8); this tests whether Meta experts must
be shared across the full depth or can instead be shared within smaller
depth-local groups. The combined point F2L2$+$Loop2 applies both topological
modifications.

The \emph{pure-Meta} family removes Local experts entirely and
routes the full top-$8$ from a single Meta pool shared across all layers
($128$ per layer, $128\times L$ in total). It constitutes the $100\%$-meta end of
the ratio axis and, like every other variant, preserves the $128\times L$ budget
and the top-$8$ activation.

\paragraph{Param-reduction ablation.}
The preceding variants all fix the routed-expert budget at $128\times L$; the
ratio family, for example, reallocates experts between meta and local within each
layer's $128$ but leaves the total unchanged. This ablation instead probes how
much shared Meta capacity is required, allowing the total to fall below
$128\times L$. Starting from the F2L2$+$Loop2 point, we reduce the per-layer Meta
width from the r75 value of $96$ through $\{90,84,72,60,48\}$ \emph{without}
refilling the freed slots with Local experts. We denote each setting
\texttt{mpl}$M$, for \textbf{m}eta-\textbf{p}er-\textbf{l}ayer width $M$
(Table~\ref{tab:ladder-mpl}).

Because only the shared Meta pool is reduced and Local is not refilled, the total
routed-expert count---and hence the stored (non-embedding routed) parameters---%
falls below the $128\times L$ budget, while the per-token routing remains at
top-$8$ ($4$ meta $+\,4$ local), leaving the activated experts and the training
FLOPs unchanged. The sweep thus isolates a single variable, shared Meta capacity,
against loss at fixed compute. It is run at all four depths, each on its standard
per-depth schedule (Table~\ref{tab:ladder-compute}), so the comparison remains
iso-compute within every rung.

% -----------------------------------------------------------------------------
% Table: partial-Meta variant sweep (per-layer topology; all depths 12/16/20/24)
% -----------------------------------------------------------------------------
\begin{table}[t]
  \centering
  \small
  \setlength{\tabcolsep}{4.5pt}
  \renewcommand{\arraystretch}{1.15}
  \caption{The Meta Experts variants. Each row is one variant,
    run at \emph{all four} depths $L\in\{12,16,20,24\}$; only the routed-expert
    layers differ from the vanilla backbone of Table~\ref{tab:ladder-arch}, and
    every variant keeps the $128\times L$ routed-expert budget (iso-param) and
    the per-token top-$8$ activation (iso-FLOP). ``Meta/layer'' denotes the per-layer-equivalent allocation used to construct the shared cross-layer Meta-expert pool, while ``Local'' denotes the number of layer-private Local experts per layer; topk lists meta$+$local experts activated per token.
    ``Local-only layers'' names the boundary layers restricted to $128$ Local
    experts (no meta); ``Meta loop'' is the number of interleaved sub-pools the
    shared Meta pool is looped through. The last column is the label used in the
    main text.}
  \label{tab:ladder-variants}
  \begin{tabular}{llccccl}
    \toprule
    Family & Variant & Meta/layer & Local & Topk (m$+$l) & Local-only & Meta \\
           &         & (share)    &       &              & layers     & loop \\
    \midrule
    Vanilla        & ---      & --- (0\%)    & 128 & 0$+$8 & ---            & 1 \\
    \midrule
    Ratio          & MR25     & 32 (25\%)    & 96  & 4$+$4 & ---            & 1 \\
                   & MR50     & 64 (50\%)    & 64  & 4$+$4 & ---            & 1 \\
                   & MR75     & 96 (75\%)    & 32  & 4$+$4 & ---            & 1 \\
    \midrule
    First/last     & F1L1     & 96 (75\%)    & 32  & 4$+$4 & first/last $1$ & 1 \\
    local-only     & F2L2     & 96 (75\%)    & 32  & 4$+$4 & first/last $2$ & 1 \\
                   & F4L4     & 96 (75\%)    & 32  & 4$+$4 & first/last $4$ & 1 \\
    \midrule
    Meta loop      & Loop2    & 96 (75\%)    & 32  & 4$+$4 & ---            & 2 \\
                   & Loop4    & 96 (75\%)    & 32  & 4$+$4 & ---            & 4 \\
                   & Loop8    & 96 (75\%)    & 32  & 4$+$4 & ---            & 8 \\
    \midrule
    Combined       & F2L2$+$Loop2 & 96 (75\%) & 32 & 4$+$4 & first/last $2$ & 2 \\
    \midrule
    Pure-Meta      & Meta-only & 128 (100\%) & 0  & 8$+$0 & ---            & 1 \\
    \bottomrule
  \end{tabular}
\end{table}

% -----------------------------------------------------------------------------
% Table: shared-Meta param-reduction ablation off F2L2+Loop2 (all depths)
% -----------------------------------------------------------------------------
\begin{table}[t]
  \centering
  \small
  \setlength{\tabcolsep}{6pt}
  \renewcommand{\arraystretch}{1.15}
  \caption{Shared-Meta param-reduction ablation, run at all four depths off the
    F2L2$+$Loop2 point. Only the per-layer Meta width is shrunk from the r75 base
    of $96$; Local ($32$) and the per-token top-$8$ ($4$ meta $+\,4$ local) are
    unchanged, so the activated FLOPs are fixed while the total routed-expert
    count (stored parameters) falls below the $128L$ budget. The shared Meta pool
    is $\mathrm{Meta/layer}\times(L-4)$ (the four boundary layers are Local-only,
    so they carry no meta). The $L{=}16$ pool size is shown as an example.}
  \label{tab:ladder-mpl}
  \begin{tabular}{lccccc}
    \toprule
    Variant       & Meta/layer & Local & Topk (m$+$l) & Meta loop & Meta pool ($L{=}16$) \\
    \midrule
    mpl90         & 90 & 32 & 4$+$4 & 2 & $1080$ \\
    mpl84         & 84 & 32 & 4$+$4 & 2 & $1008$ \\
    mpl72         & 72 & 32 & 4$+$4 & 2 & $864$ \\
    mpl60         & 60 & 32 & 4$+$4 & 2 & $720$ \\
    mpl48         & 48 & 32 & 4$+$4 & 2 & $576$ \\
    \bottomrule
  \end{tabular}
\end{table}

\subsection{Discussion for the Residual Analysis.}
\label{appendix:discussion-residual}

Two remarks keep the claim honest. First, the path through the shared pool
has $O(1)$ length---one hop from layer $l$ into $\theta$ and one hop out to
layer $m$---independent of the layer distance $|m-l|$, whereas both
pre-existing routes of layer-wise MoE (forward carrying and the gradient
highway) pass through $|m-l|$ Jacobians. The effective local gain is
governed by the spectrum of the cross-layer tangent factor
$J_m^{(t)}{J_l^{(t)}}^\top$ appearing in
Eq.~\eqref{eq:kernel}; the analysis therefore characterizes access and
path length, rather than guaranteeing against vanishing signals.
Second, the causal direction of a single forward pass is unchanged---
$\hat{x}_m$ does not depend on the activations of deeper layers within the
same pass; what a shallow layer accesses is the \emph{content written into
the shared pool by deeper layers during training}. This is exactly
analogous to how any trained weight carries information about the data it
was trained on, with the difference that here the carrier is shared across
depth, making the information accessible at every layer in a common code
space.

Finally, layer-wise gradient reweighting,
$\bar{\theta} = \sum_l \alpha_l g_l$
(e.g., norm equalization or conflict projection), amounts to gating this
backward pathway: it modulates how much of each layer's write
$\Delta\theta_l^{(t)}$ is allowed to influence the other layers' reads
through $\Delta y_{m\leftarrow l}^{(t)}$, and hence through the accumulated
cross-layer influence $\mathcal{I}_{m\leftarrow l}$.

\end{document}